\documentclass{article}
\usepackage{spconf,amsmath,epsfig}

\usepackage{setspace}
\usepackage{epstopdf}
\usepackage{amsmath,amsfonts,amssymb}
\usepackage{bbm}
\usepackage{graphicx}
\usepackage[colorlinks=true, allcolors=blue]{hyperref}
\usepackage{tikz}
\usepackage{verbatim}
\usepackage{siunitx}
\usepackage{graphics}
\usepackage{subfig}
\usepackage{multirow}
\usepackage{multicol}
\usepackage{xspace}
\usepackage{colortbl}
\usepackage[acronym]{glossaries}

\title{ASSESSMENT OF SENTINEL-2 SPATIAL AND TEMPORAL COVERAGE BASED ON THE SCENE CLASSIFICATION LAYER}
\name{
\begin{tabular}{@{}c@{}}
Cristhian Sanchez$^{1, 2}$ \qquad Francisco Mena$^{1, 2}$ \\ 
Marcela Charfuelan$^{2}$ \qquad Marlon Nuske$^{2}$ \qquad Andreas Dengel$^{1, 2}$
\end{tabular}
}
\address{$^{1,}$University of Kaiserslautern-Landau (RPTU), Kaiserslautern, Germany\\
     $^{2}$German Research Center for Artificial Intelligence (DFKI), Kaiserslautern, Germany\\
     }

\newcommand{\cloudfilter}{L-all-but-cloud\xspace}
\newcommand{\vegfilter}{L-veg-non-veg\xspace}

\newacronym{s2}{S2}{Sentinel-2}
\newacronym{ml}{ML}{machine learning}
\newacronym{scl}{SCL}{scene classification layer}
\newacronym{sits}{SITS}{satellite image time series}
\newacronym{si}{SI}{satellite image}
\newacronym{rs}{RS}{remote sensing}
\newacronym{rf}{RF}{random forest}
\newacronym{sca}{SCA}{spatial coverage assessment}
\newacronym{tca}{TCA}{temporal coverage assessment}

\begin{document}
%
\maketitle
\begin{abstract}
Since the launch of the \gls{s2} satellites, many ML models have used the data for diverse applications.
The \gls{scl} inside the \gls{s2} product provides rich information for training, such as filtering images with high cloud coverage.
However, there is more potential in this. 
We propose a technique to assess the clean optical coverage of a region, expressed by a SITS and calculated with the \gls{s2}-based \gls{scl} data.
With a manual threshold and specific labels in the \gls{scl}, the proposed technique assigns a percentage of spatial and temporal coverage across the time series and a high/low assessment. 
By evaluating the AI4EO challenge for Enhanced Agriculture, we show that the assessment is correlated to the predictive results of ML models. The classification results in a region with low spatial and temporal coverage is worse than in a region with high coverage.
Finally, we applied the technique across all continents of the global dataset LandCoverNet. 
\end{abstract}

\vspace{0.2cm}
\textit{Copyright 2024 IEEE. Published in the 2024 IEEE International Geoscience and Remote Sensing Symposium (IGARSS 2024), scheduled for 7 - 12 July, 2024 in Athens, Greece. Personal use of this material is permitted. However, permission to reprint/republish this material for advertising or promotional purposes or for creating new collective works for resale or redistribution to servers or lists, or to reuse any copyrighted component of this work in other works, must be obtained from the IEEE. Contact: Manager, Copyrights and Permissions / IEEE Service Center / 445 Hoes Lane / P.O. Box 1331 / Piscataway, NJ 08855-1331, USA. Telephone: + Intl. 908-562-3966.
}

\section{INTRODUCTION} \label{sec:intro}  
Optical observations are the fuel of many \gls{rs}-based applications. These images rely on passive observation that is affected by distinct factors, such as clouds, haze, cloud shadow, and snow \cite{shen2015missing}. 

\gls{s2}-based optical data has played a key role in different research fields related to land cover-use mapping with \gls{ml} over the last decade. 
The inclusion of the \gls{scl} has been crucial for
filtering images with a high presence of clouds \cite{hardy2021sen2grass,tarasiewicz2022extracting}. The \gls{scl} is a \gls{s2} product that provides an estimated scene class for each pixel in the paired \gls{s2} image, at a 10-meter pixel resolution. In this work, we present an assessment based on the information contained in the \gls{scl} data.
The research question that drags us is {\it how much clean data, e.g. cloud-free, in a time series are we actually feeding into \gls{ml} models?}, first thoughts brought us to inspect the cloud coverage per each satellite image in a time-series.
However, we decided to present a general assessment based on user-defined labels in the \gls{scl}. 
Our proposed assessment calculates the spatial and temporal coverage in a sample region expressed by \gls{sits}, concretely, the \gls{scl} inside the \gls{s2} data.

We evaluate the relation between the spatial and temporal coverage with the classification results of a \gls{rf} trained in the AI4EO Enhanced Agriculture dataset. The \gls{rf} model is trained in a pixel-wise manner with neighborhood information \cite{tarasiewicz2022extracting}.
We obtained that the classification results are worse in sample regions with cloudy conditions.
In addition, we calculated the spatial and temporal coverage on the recent LandCoverNet global dataset \cite{alemohammad2020landcovernet} and obtained a distribution of clean coverage per continent in the year 2018 based on the \gls{s2} data.
This coverage calculation could be useful for researchers interested in assessing the quality of \gls{sits} and understanding the prediction differences by region.
We provide an evaluation with the \gls{scl} inside the \gls{s2} data, but this could be reproduced for any other scene classification mask. 
The code and assessment obtained can be found at \url{https://github.com/fmenat/SITS_S2Coverage}. 

This paper is organized as follows. The background is presented in Sec.~\ref{sec:works}, followed by the proposed assessment in Sec.~\ref{sec:proposal}. In Sec.~\ref{sec:evaluation}, we show the results of the assessment in two datasets. 
Finally, Sec.~\ref{sec:conclu} provides the conclusion of our work.


\begin{table*}[!t]
\begin{minipage}{0.24\textwidth}
    \small
    \centering
    \begin{tabular}{c|c} \hline
        Tag & Name \\ \hline
        0 & No Data \\
        1 & Saturated or Defective \\
        2 & Dark Area Pixels \\
        3 & Cloud Shadows \\ 
        4 & Vegetation \\
        5 & Not Vegetated \\
        6 & Water \\ 
        7 & Unclassified \\
        8 & Cloud Medium \\
        9 & Cloud High \\
        10 & Thin Cirrus \\
        11 & Snow \\ \hline
    \end{tabular}
    \caption{Possible labels in the \gls{s2}-based \gls{scl} data.} \label{tab:scl}
\end{minipage}
\hfill
\begin{minipage}{0.74\textwidth}
    \centering
    \includegraphics[width=\linewidth]{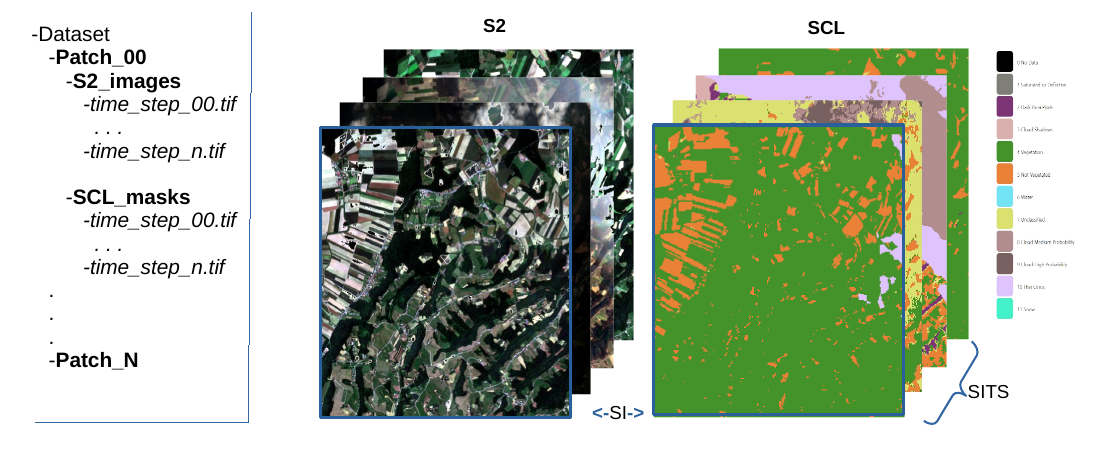}\\%
    \caption{Structure example of the optical \gls{sits} with the corresponding \gls{scl} time series mask.}\label{fig:method:data}%
\end{minipage}
\end{table*}

\section{BACKGROUND AND RELATED WORK} \label{sec:works}

Thanks to the European Space Agency and the Copernicus \gls{s2} mission, it is possible to access optical satellite image data. 
The \Gls{s2} is a constellation of two sun-synchronized satellites that are placed along the same orbit and separated by 180 degrees. This allows access to sun-lighted images from the same location on Earth at approximately every five days.  \gls{s2}-based optical images are composed by 13 bands with a spatial resolution varying from 10 to 60 meters. 
Along the optical images, a \gls{scl} generated by the Sen2Cor algorithm \cite{main2017sen2cor} is provided. The main purpose of this layer is to deliver a close estimation of what is presented in each pixel of the paired optical image. 
The available classes are shown in Table \ref{tab:scl}. 

There are different factors that affect the clean observation of optical images \cite{shen2015missing}, such as cloud, haze, snow, anomalies, and errors. 
Indeed, some works have shown that \gls{ml} models trained on optical data get low predictive performance in cloudy conditions \cite{garnot2022multi,ferrari2023fusing}. 
Ferrari et al. \cite{ferrari2023fusing} categorize regions into three different cloud coverage conditions (low, medium, and high) for deforestation prediction with optical and radar \gls{sits}. 
Nevertheless, in this manuscript, we generalize previous analysis beyond cloud presence.

Consequently, the \gls{scl} has been used in multiple studies to discriminate (filter out) data belonging to cloud-related labels \cite{hardy2021sen2grass}, water and snow \cite{roteta2019development}, or defective pixels \cite{johnson2022opensentinelmap}. Furthermore, another usage is the selection of pixels belonging to a certain class. 
In AI4Boundaries work \cite{d2023ai4boundaries}, the tags 2, 4, 5, 6 and 7 in the \gls{scl} (See Tab.~\ref{tab:scl}) are considered as \textit{clean} input data for the \gls{ml} training.


\section{Assessment of Coverage Availability} \label{sec:proposal}
Regardless of the spatial and temporal resolutions of satellite images, we consider the following terminology. 
\textit{Sample region}, as a single sample zone that is used to define a region of interest. 
The data in a sample region could come from a single \gls{si}, or a \gls{sits} as a (ordered) collection of images at different times.
\textit{Clean coverage}, which refers to the spatio-temporal availability of data pixels belonging to specific classes. 
See Figure~\ref{fig:method:data} for an illustration. 
In addition, consider for each sample region $i$ an optical SITS $\mathcal{X}^{(i)}$, (with $B$ bands) and its corresponding SCL data (also a SITS), $\mathcal{L}^{(i)}$. Both information with a pixel resolution of $W\times H$:
\begin{align*}
    \mathcal{X}^{(i)} &= \left\{ X_1^{(i)},  X_2^{(i)}, \ldots,  X_{T_i}^{(i)} \right\},  \ \text{where } X_{t,w,h}^{(i)} \in \mathbb{R}_{+}^{ B} \\
    \mathcal{L}^{(i)} &= \left\{  L_1^{(i)}, L_2^{(i)}, \ldots,  L_{T_i}^{(i)} \right\}, \  \text{where }  L_{t,w,h}^{(i)} \in \left[0,11 \right] 
\end{align*}

Given a set of labels $\mathcal{K}$, we define the spatial coverage $SC_i^{(t)}$ for a \gls{si} at time-step $t$ in the sample region $i$ as the percentage of pixels in $L_t^{(i)}$ belonging to the set $\mathcal{K}$. Besides, we define the spatial coverage $SC^{(i)}$ for the whole \gls{sits} in a sample region $i$, as the average across the time-series:
\begin{align}
    SC_t^{(i)} &= \frac{1}{W \cdot H} \sum_{w}^W \sum_{h}^H \mathbbm{1}(L_{t,w,h}^{(i)} \in \mathcal{K} ) \\
    SC^{(i)} & = \frac{1}{T_i} \sum_{t=1}^{T_i} SC_t^{(i)}\ ,
\end{align}
with $\mathbbm{1}$ a function giving $1$ when the equation inside holds.
Then, considering a threshold $SC_{\text{thresh}}$, we define a \gls{sca} label for each sample region $i$,
\begin{equation}
    \text{SCA}^{(i)} = \left\{ \begin{array}{ll}
              \text{high} & \text{if}\ SC^{(i)} \geq SC_{\text{thresh}}\\
              \text{low}  & \text{otherwise.}
             \end{array} \right.
\end{equation}
In addition, we define the temporal coverage $TC^{(i)}$ in a sample region $i$, based on the spatial coverage across the time-series.
Based on a threshold $TC_{\text{thresh}}$, we defined a \gls{tca} label for each sample region $i$,
\begin{align}
    TC^{(i)} &= \frac{1}{T_i} \sum_{t=1}^{T_i} \mathbbm{1}( SC_t^{(i)} \geq TC_{\text{thresh}} ) \\
    \text{TCA}^{(i)} &= \left\{ \begin{array}{ll}
              \text{high} & \text{if}\ TC^{(i)} \geq TC_{\text{thresh}}\\
              \text{low}  & \text{otherwise.}
             \end{array} \right.
\end{align}
Therefore, given the \gls{scl}-based \gls{sits} as input data, $\mathcal{L}^{(i)}$, our technique obtains: $SC^{(i)}$, $TC^{(i)}$, $\text{SCA}^{(i)}$, and $\text{TCA}^{(i)}$.
The final goal of the proposed technique is to rapidly get an assessment of the spatio-temporal availability in a region.

\section{EVALUATION AND APPLICATION} \label{sec:evaluation}

\subsection{Evaluation: AI4EO Enhanced Agriculture} \label{sec:evaluation:ai4eo}
\begin{figure*}[!t]
    \centering
    \includegraphics[width=0.75\linewidth]{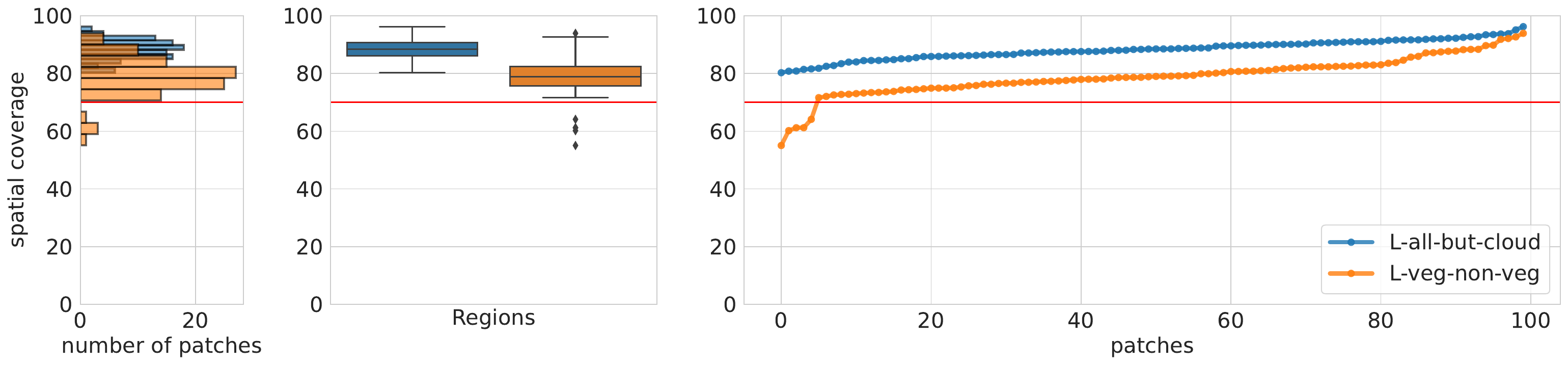}\\%
    \includegraphics[width=0.75\linewidth]{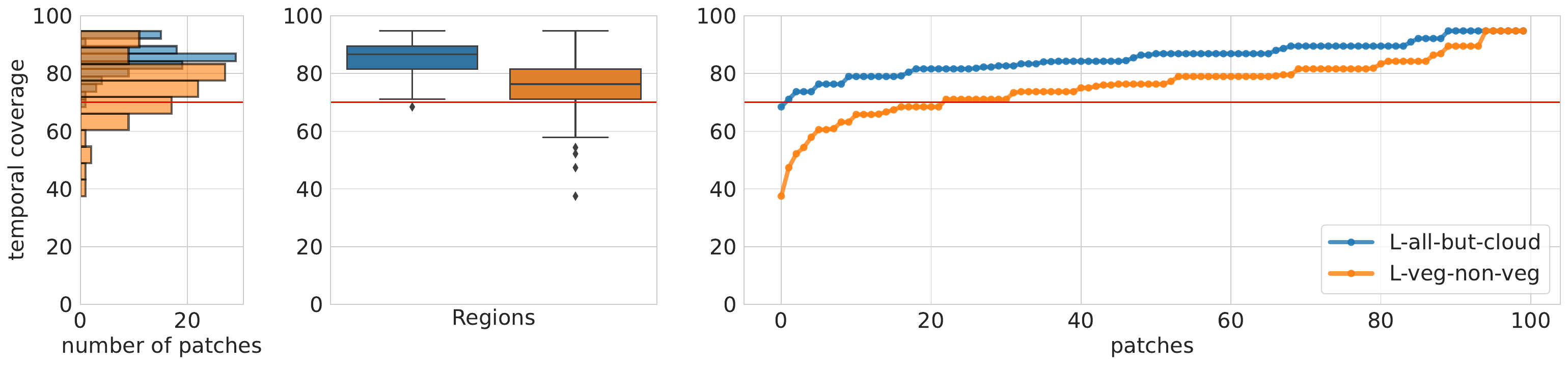}%
    \caption{Spatial and temporal coverage in the AI4EO Enhanced Agriculture dataset. Two types of filters were used: \cloudfilter as a cloud-removal filtering, \vegfilter as a vegetated-related filtering. A 70\% coverage is shown in red.}\label{fig:ai4eo_coverage}%
\end{figure*} 
We evaluated the assessment in the AI4EO challenge on Enhanced Agriculture\footnote{\url{platform.ai4eo.eu/enhanced-sentinel2-agriculture}}. The input data consists of \gls{s2}-based optical \gls{sits} across the growing season in 2019 for Slovenia (March to September). The target data consists of a binary masks (cultivated or not) at a higher spatial resolution (2.5 m) than the \gls{s2}. The (100) sample regions are $500\times 500$ size SITS of variable length.
First, we calculate the spatial and temporal coverage of the sample regions by considering two types of filters. \cloudfilter, that represents a cloud-removal filter, where all the classes are selected for coverage except cloud-related (3, 8 and 9). On the other hand, \vegfilter that represents a vegetated-related filter, where only classes 4 and 5 are selected for coverage. 
Figure~\ref{fig:ai4eo_coverage} shows the coverage for these two filters. It can be seen that the \vegfilter coverage is lower than \cloudfilter coverage, since it filters more label-types for its calculation. 

For prediction, we used the approach by Tarasiewicz et al. \cite{tarasiewicz2022extracting}. First, a bilinear interpolation is carried out to match the input data to the target data. Then, a statistic generation is performed for each input pixel on a neighborhood of $25\times 25$. Finally, a \gls{rf} model it uses to predict the binary label of the central pixel based on the pixel neighborhood time series. The training is performed only on the central pixel from the original 10 meters resolution image \cite{tarasiewicz2022extracting}. We use 80 sample regions for training and 20 for validation, and evaluate with the same metrics used in the challenge: Matthews Correlation Coefficient (MCC) and Accuracy (ACC).

\begin{figure}[!t]
    \centering
    \includegraphics[width=\linewidth]{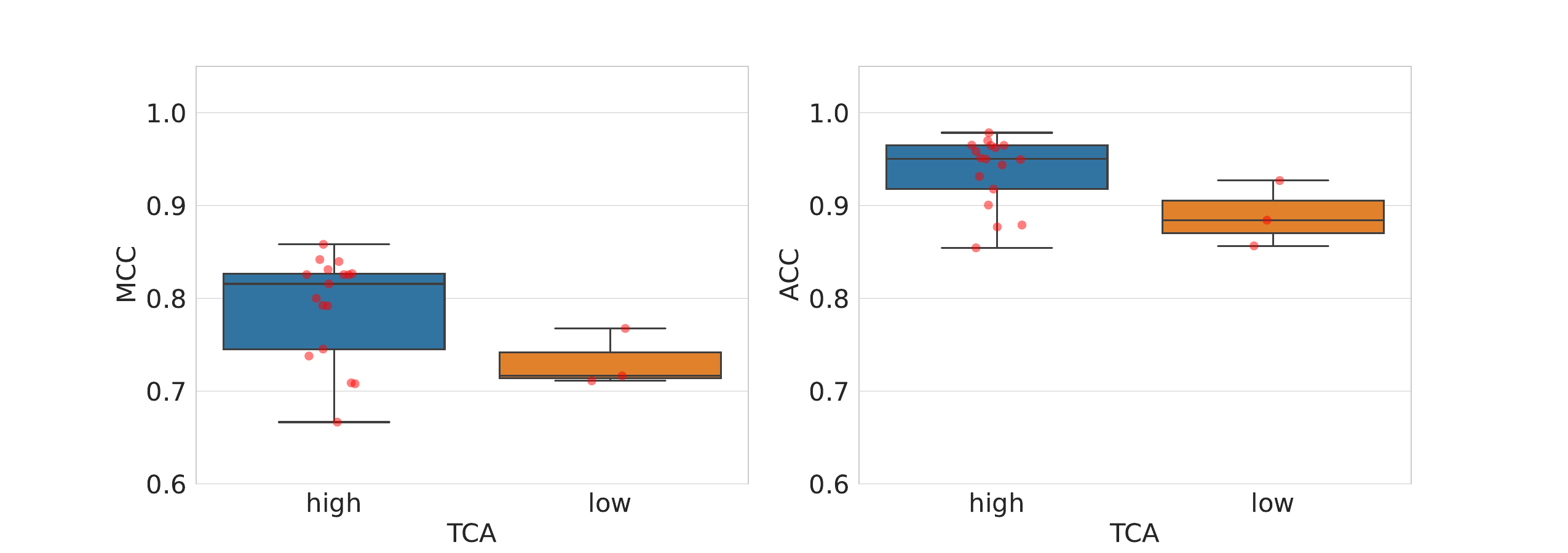}%
    \caption{Classification results of \gls{rf} in sample regions categorized as \textit{high} and \textit{low} by the \gls{tca} with a 70\% threshold. Each point represents the averaged metric on a specific sample region. The \vegfilter criteria is used for the assesment.} \label{fig:example}%
\end{figure}
Figure~\ref{fig:example} shows the classification results of sample regions by the \vegfilter (since with the \cloudfilter we obtained only one field categorized as low). It is clear that in regions with low spatial and temporal coverage, the classification results are worse than in high coverage. 
\begin{figure}[!t]
    \centering
    \includegraphics[width=\linewidth]{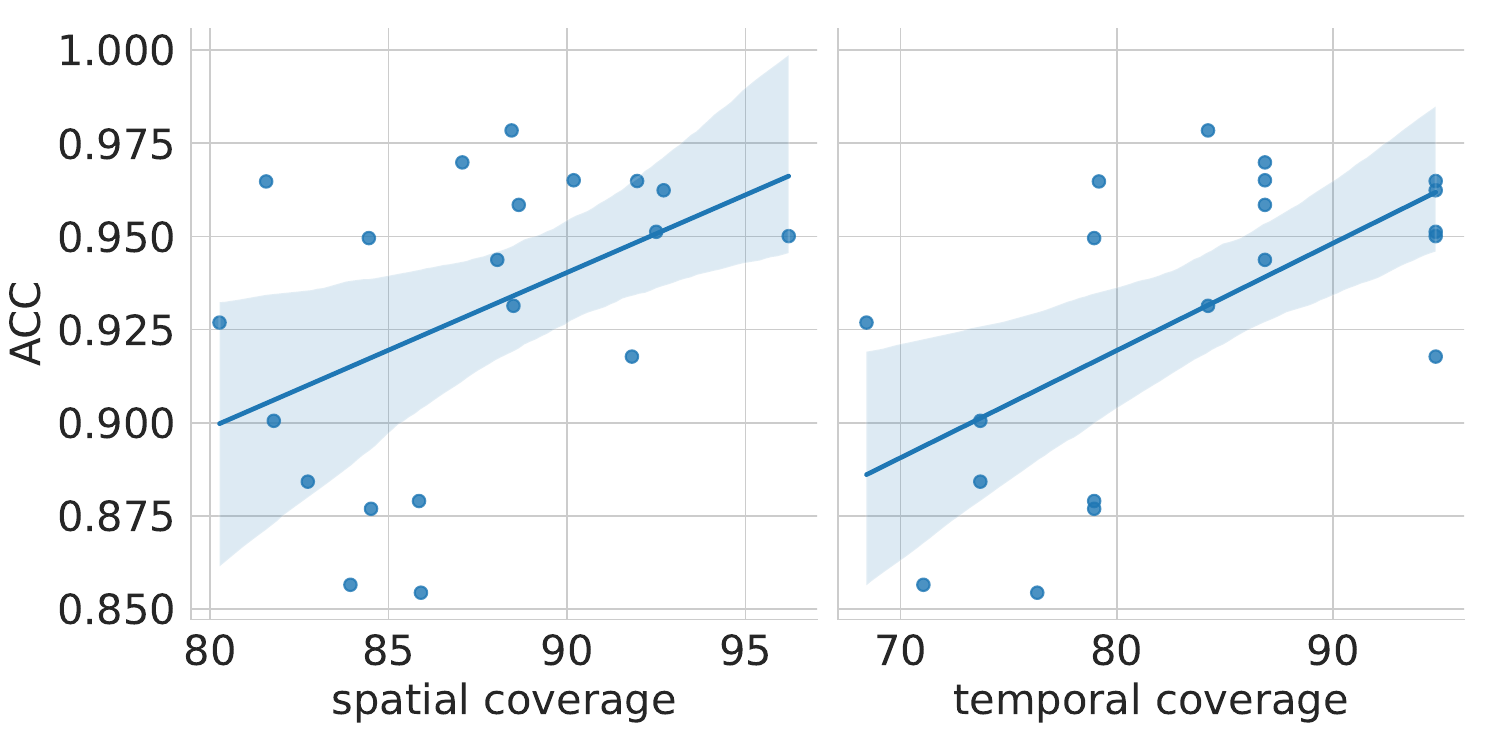}%
    \caption{Accuracy of \gls{rf} in different sample regions based on \cloudfilter. Each point represents the averaged metric on a specific sample region. The correlation of ACC with spatial coverage is $45.0$ and with temporal coverage is $70.3$.} \label{fig:example_corr}%
\end{figure}
Furthermore, the Figure~\ref{fig:example_corr} illustrates the correlation between the accuracy and the spatio-temporal coverage with the \cloudfilter filter. There is a tendency that, when a sample region has a higher temporal coverage, the classification is better.

\subsection{Application: LandCoverNet}  \label{sec:evaluation:landcovernet}
\begin{figure*}[!t]
    \centering
    {\includegraphics[width=0.75\linewidth]{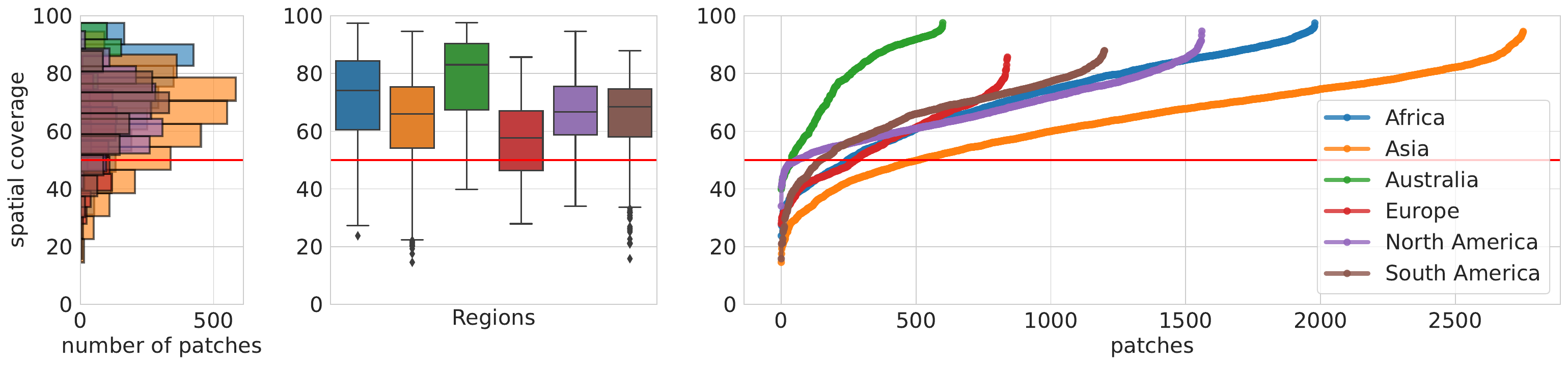}}\\%
    {\includegraphics[width=0.75\linewidth]{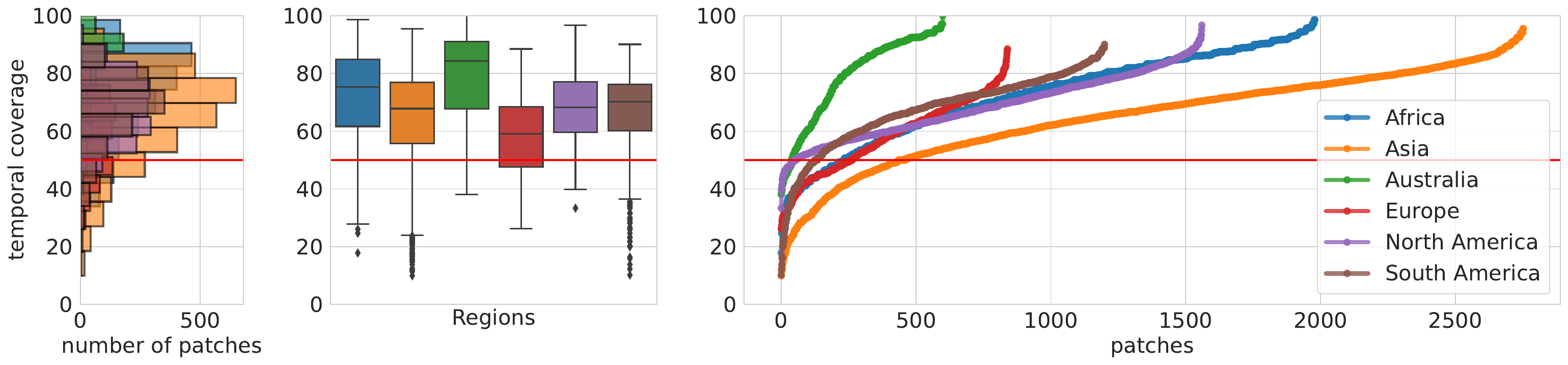}}\\
    \caption{Spatial and temporal coverage in the LandCoverNet with the \cloudfilter criteria. A 50\% coverage is shown in red.}\label{fig:landcovernet_coverage}%
\end{figure*}
In addition, we applied the technique to the recent global dataset of LandCoverNet \cite{alemohammad2020landcovernet}.
There are sample regions coming from all the continents: 1980 in Africa, 2753 in Asia, 600 in Australia, 840 in Europe, 1561/1200 in North/South America. 
A \gls{s2}-based \gls{sits} (optical and \gls{scl} data) is available across 2018 for each sample region.
These regions correspond to a $256\times 256$ size SITS of variable length.

Figure~\ref{fig:landcovernet_coverage} shows the spatial and temporal coverage with the \cloudfilter filter on each continent of the LandCoverNet dataset. 
As expected, it can be seen that each continent has different clean coverage patterns, with Australia having on average a higher cloudless coverage in 2018. In contrast, Europe and Asia are the regions with more cloudy conditions on average.
In addition, the coverage distribution across sample regions within each continent is quite different. With a threshold of 50\%, the number of samples regions categorized with low \gls{tca} are: 230 in Africa, 435 in Asia, 38 in Australia, 255 in Europe, 46 in North America, and 124 in South America.
However, as each continent has a different total number of sample regions, the percentage with low \gls{tca} is 12\% in Africa, 16\% in Asia, 5\% in Australia,  30\% in Europe, 3\% in North America, and 15\% in South America.
While Asia has almost twice of low \gls{tca} sample regions than Africa and Europe, this is a low value relative to the total number of sample regions in each continent.
Figure~\ref{fig:landcovernet_coverage} also shows some outlier sample regions with very low coverage in different continents, with Asia and South America the more clear cases. 
Surprisingly, Australia and Europe do not have outlier sample regions regarding the spatial and temporal coverage.


\section{CONCLUSION} \label{sec:conclu}
We proposed a technique to assess the amount of clean coverage in sample regions from \gls{s2}-based \gls{sits}. 
The purpose is to assess the spatio-temporal clean coverage in a region of interest based on the \gls{scl}-based \gls{sits} contained in the \gls{s2} product.
Our evaluation shows a positive correlation between the clean coverage and the predictive results of \gls{ml} models trained with \gls{s2}-based \gls{sits}. 
Some potential directions for this research could be curriculum learning. For instance, provide an order of sample regions during training from the highest to lowest coverage sample regions or vice versa.

\textbf{Acknowledgement}. F. Mena acknowledges the financial support from the chair of Prof. A. Dengel with RPTU.

\small
\bibliographystyle{IEEEbib}
\bibliography{refs}

\end{document}